\documentclass[letterpaper, 10 pt, conference]{styles/ieeeconf}
\IEEEoverridecommandlockouts                             
\usepackage{url}
\usepackage{color}
\usepackage{xcolor}
\usepackage{wrapfig}
\newif\iffigs
\figstrue 

\usepackage[fleqn]{amsmath}
\usepackage{float}
\usepackage{optidef}
\usepackage{setspace}
\usepackage{graphicx}
\usepackage{mathrsfs}
\usepackage{amssymb}
\usepackage{nicefrac}
\usepackage{algorithm}
\usepackage[noend]{algpseudocode}

\usepackage{flushend}
\usepackage{multicol}

\makeatletter
\newcommand\fs@spaceruled{\def\@fs@cfont{\bfseries}\let\@fs@capt\floatc@ruled
  \def\@fs@pre{\vspace{0.4\baselineskip}\hrule height.8pt depth0pt \kern2pt}%
  \def\@fs@post{\vspace{-0.4\baselineskip}\kern2pt\hrule\relax\vspace{-12pt}}%
  \def\@fs@mid{\kern2pt\hrule\kern2pt}%
  \let\@fs@iftopcapt\iftrue}
\makeatother

\usepackage{lipsum}
\usepackage{pgfplots}
\pgfplotsset{
    tick label style={font=\bfseries\color{white!15!black}},
}
\pgfplotsset{compat=newest}
\newlength\figH
\newlength\figW

\DeclareMathOperator*{\argmin}{arg\,min}
\DeclareMathOperator{\proj}{proj}

\usepackage{hyperref}

\definecolor{melon}{rgb}{0.99, 0.74, 0.71}
\definecolor{my_gray}{HTML}{616B85}

\IEEEoverridecommandlockouts

\title{\LARGE \bf TinyMPC: Model-Predictive Control on \\ Resource-Constrained Microcontrollers}

\author{Anoushka Alavilli*, Khai Nguyen*, Sam Schoedel*, Brian Plancher, Zachary Manchester
\thanks{$*$These authors contributed equally and are listed in alphabetical order.}
\thanks{Anoushka Alavilli is with the Department of Electrical \& Computer Engineering, Carnegie Mellon University, 5000 Forbes Ave., Pittsburgh, PA 15213. {\tt\footnotesize apalavil@andrew.cmu.edu}}%
\thanks{Khai Nguyen is with the Department of Mechanical Engineering, Carnegie Mellon University, 5000 Forbes Ave., Pittsburgh, PA 15213. {\tt\footnotesize xuankhan@andrew.cmu.edu}}%
\thanks{Samuel Schoedel and Zachary Manchester are with the Robotics Institute, Carnegie Mellon University, 5000 Forbes Ave., Pittsburgh, PA 15213. {\tt\footnotesize \{sschoede, zmanches\}@andrew.cmu.edu}}%
\thanks{Brian Plancher is with Barnard College, Columbia University, 3009 Broadway New York, NY 10027. {\tt\footnotesize bplancher@barnard.edu}}%
}
\usepackage{amsmath, amssymb}
\begin{document}
\maketitle
\thispagestyle{empty}
\pagestyle{empty}


\begin{abstract}
    Model-predictive control (MPC) is a powerful tool for controlling highly dynamic robotic systems subject to complex constraints. However, MPC is computationally demanding, and is often impractical to implement on small, resource-constrained robotic platforms. We present TinyMPC, a high-speed MPC solver with a low memory footprint targeting the microcontrollers common on small robots. Our approach is based on the alternating direction method of multipliers (ADMM) and leverages the structure of the MPC problem for efficiency. We demonstrate TinyMPC's effectiveness by benchmarking against the state-of-the-art solver OSQP, achieving nearly an order of magnitude speed increase, as well as through hardware experiments on a 27 gram quadrotor, demonstrating high-speed trajectory tracking and dynamic obstacle avoidance. TinyMPC is publicly available at {\color{blue}{\url{https://tinympc.org}}}.
\end{abstract}

\section{Introduction} \label{sec:intro}

Model-predictive control (MPC) enables reactive and dynamic online control for robots while respecting complex control and state constraints such as those encountered during dynamic obstacle avoidance and contact events~\cite{wensing2022optimization,di2020software,manchester2019contact,Kuindersma23Talk}. However, despite MPC's many successes, its practical application is often hindered by computational limitations, which can necessitate algorithmic simplifications~\cite{plancher2020performance,Neuman21}. This challenge is amplified when dealing with systems that have fast or unstable open-loop dynamics, where high control rates are needed for safe and effective operation.

At the same time, there has been an explosion of interest in tiny, low-cost robots that can operate in confined spaces, making them a promising solution for applications ranging from emergency search and rescue~\cite{mcguire2019minimal} to routine monitoring and maintenance of infrastructure and equipment~\cite{de2018inverted,duisterhof2021sniffy}. These robots are limited to low-power, resource-constrained microcontrollers (MCUs) for their computation~\cite{crazyflie_2.0,bittle}. As shown in Figure~\ref{fig:hardware}, these microcontrollers feature orders of magnitude less RAM, flash memory, and processor speed compared to the CPUs and GPUs available on larger robots and historically were not able to support the real-time execution of computationally or memory-intensive algorithms~\cite{neuman2022tiny,zhang2017visual}. Consequently, many examples in the literature of intelligent robot behaviors executed on these tiny platforms rely on off-board computers~\cite{AMSwarm,varshney2019deepcontrol,lambert2019low,luis2020online,xi2021gto,torrente2021data,chee2022knode}.

\begin{figure}[!t]
    \centering
    \includegraphics[width=0.9\columnwidth,trim={0cm 0.3cm 0 0.3cm},clip]{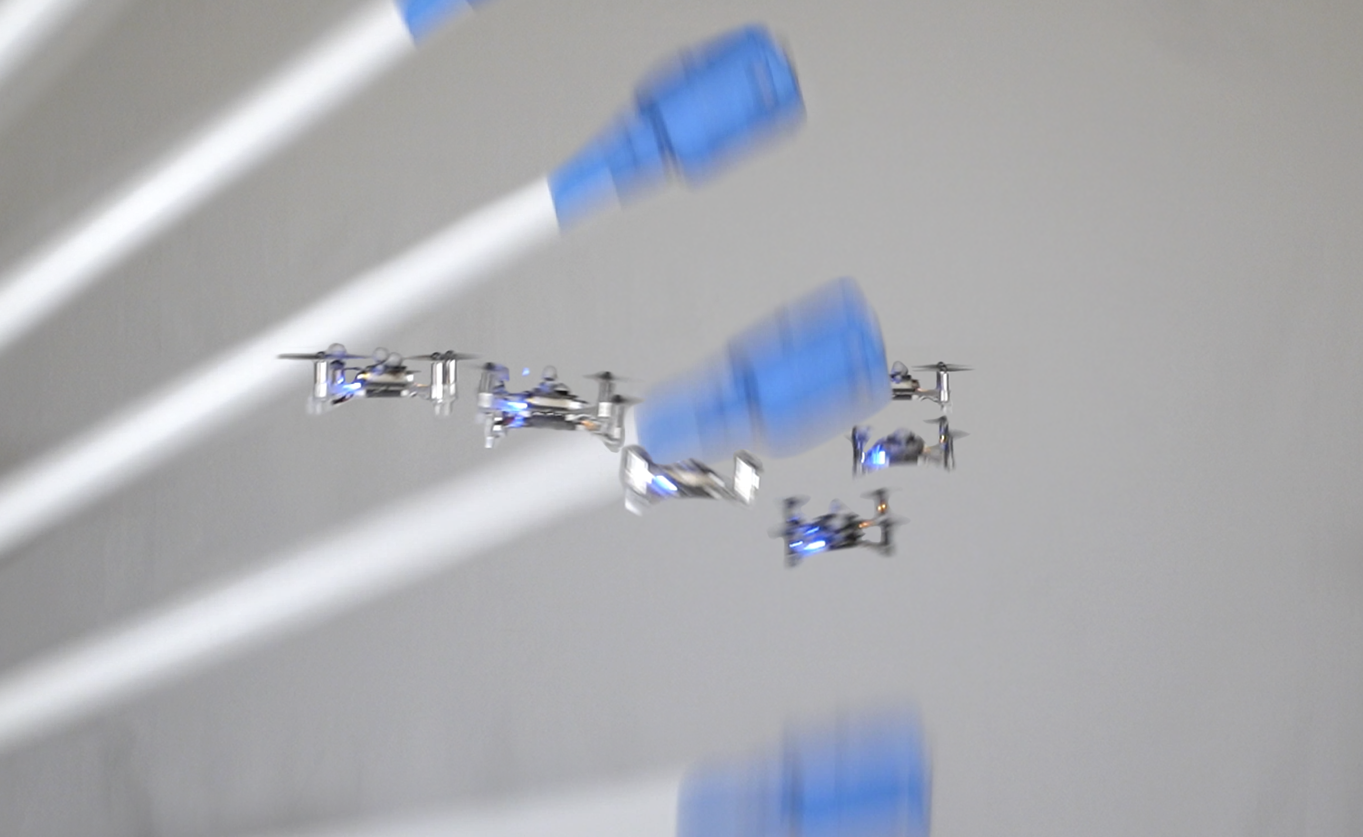}\\

    \vspace{2pt}

    \includegraphics[width=0.9\columnwidth]{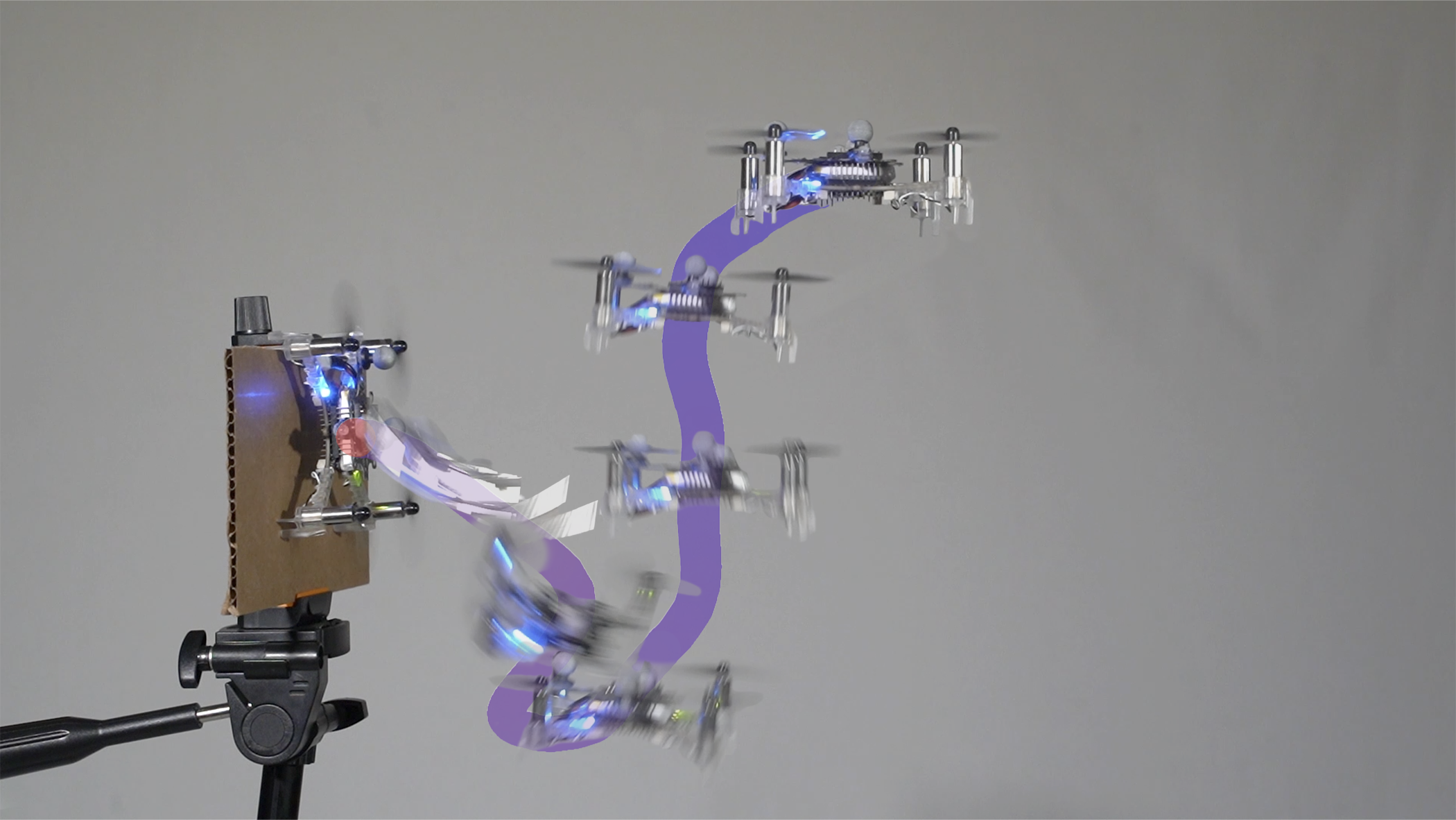}

    \caption{TinyMPC is a fast convex model-predictive control solver that enables real-time optimal control on resource-constrained microcontrollers. We demonstrate its efficacy in dynamic obstacle avoidance (top) and recovery from $90^\circ$ attitude errors (bottom) on a 27 gram Crazyflie quadrotor.}
    \label{fig:avoid2}
    \vspace{-10pt}
\end{figure}

\begin{figure*}[!t]
    \centering
    \vspace{5pt}
    \includegraphics[width=0.99\textwidth]{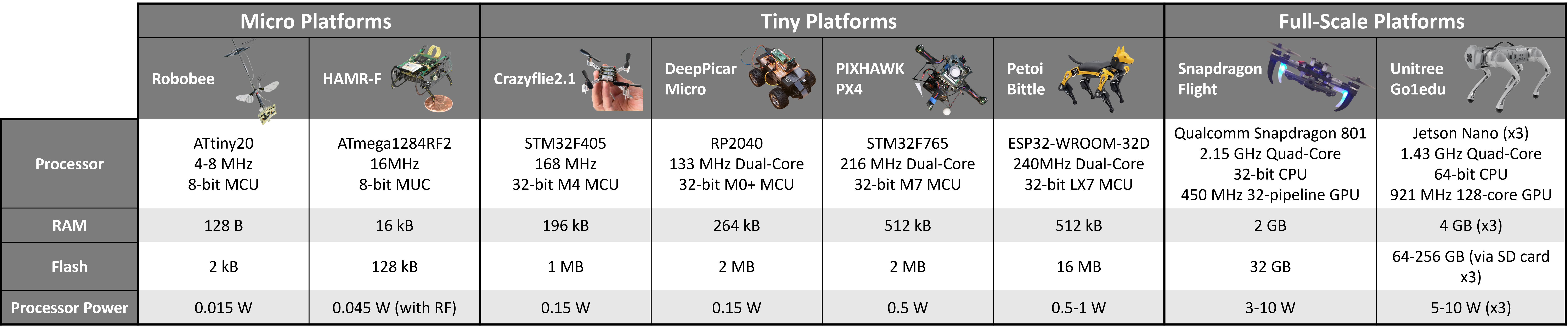}
    \caption{A comparison of micro, tiny, and full-scale robot platforms and their associated computational hardware. At the smallest scale, microrobots like the Robobee~\cite{jafferis2019untethered} and HAMR-F~\cite{goldberg2018power} use highly constrained 8-bit microcontrollers to execute pre-planned open-loop gaits or wing motions. At large scales, powerful embedded CPUs and GPUs, found onboard the Snapdragon Flight quadrotor~\cite{loianno2016estimation} or Unitree Go1edu quadruped~\cite{unitree23go1edu}, enable high performance at the cost of high power requirements. In this work we target tiny robots such as the Crazyflie2.1~\cite{crazyflie2_1}, DeepPiCarMicro~\cite{bechtel2022deeppicarmicro}, PIXHAWK PX4~\cite{meier2012pixhawk}, and Petoi Bittle~\cite{petoi23bittle} that leverage 32-bit microcontrollers for motion planning and control. These devices are capable of some onboard computation, but feature orders of magnitude less processor speed, RAM, and flash memory than full-scale robots.}
    \label{fig:hardware}
    \vspace{-10pt}
\end{figure*}

Several efficient optimization solvers and techniques suitable for embedded MPC have emerged in recent years~\cite{megahertz2014, Donoghue2013}. Notable software packages among these are OSQP~\cite{stellato2020osqp} and CVXGEN~\cite{Mattingley12}. Both of these solvers have code-generation tools that enable users to create dependency-free C code to solve quadratic programs (QPs) on embedded computers. However, they do not take full advantage of the unique structure of the MPC problem, generally making them too memory intensive and too computationally demanding to run within the resource constraints of many microcontrollers.

On the other hand, the recent success of ``TinyML'' has enabled the deployment of neural networks on microcontrollers~\cite{neuman2022tiny}. Motivated by these results, we developed TinyMPC, an MCU-optimized implementation of convex MPC using the alternating direction method of multipliers (ADMM) algorithm. At its core, our solver is designed to
    \textit{accelerate and compress} the ADMM algorithm by \textit{exploiting the structure} of the MPC problem.

In particular, we precompute and cache expensive matrix factorizations, allowing TinyMPC to completely avoid online division and matrix inversion. This approach enables rapid computation with a very small memory footprint, enabling deployment on resource-constrained MCUs. To the best of the authors' knowledge, TinyMPC is the first MPC solver tailored for execution on these MCUs that has been demonstrated onboard a highly dynamic, compute-limited robotic system. 
Our contributions include:
\begin{itemize}
    \item A novel quadratic-programming algorithm that is optimized for MPC, is matrix-inversion free, and achieves high efficiency and a very low memory footprint. This combination makes it suitable for deployment on resource-constrained microcontrollers.
    \item An open-source implementation of TinyMPC in C++ that delivers state-of-the-art real-time performance for convex MPC problems on microcontrollers.
    \item Experimental demonstrations on a small, agile, resource-constrained quadrotor platform.
\end{itemize}

This paper proceeds as follows:
Section~\ref{sec:background} reviews linear-quadratic optimal control, convex optimization, and ADMM. Section~\ref{sec:algorithm} then derives the core TinyMPC solver algorithm.
Benchmarking results and hardware experiments on a Crazyflie quadrotor are presented in Section~\ref{sec:results}. Finally, we summarize our results and conclusions in Section~\ref{sec:conclusion}.




\section{Background} \label{sec:background}
\subsection{The Linear-Quadratic Regulator}

The linear-quadratic regulator (LQR)~\cite{lewis12optimal} is a widely used approach for solving robotic control problems.
LQR optimizes a quadratic cost function subject to a set of linear dynamics constraints:
\begin{equation} \label{eq:lqr}
\begin{split}
    \min_{\substack{x_{1:N}, u_{1:N-1}}} 
        & J = \frac{1}{2}x_N^{\intercal} Q_f x_N + q_f^{\intercal} x_N + \\ 
        \sum_{k=1}^{N-1} &\frac{1}{2}x_k^{\intercal} Q x_k + q_k^{\intercal} x_k + \frac{1}{2}u_k^{\intercal} R u_k + r_k^{\intercal} u_k\\
        \textrm{subject to} \quad & x_{k+1} = A x_{k} + B u_{k} \quad \forall k \in [1,N),\\
\end{split}
\end{equation}
\noindent where $x_k \in \mathbb{R}^n$, $u_k \in \mathbb{R}^m$ are the state and control input at time step $k$, $N$ is the number of time steps (also referred to as the horizon), $A \in \mathbb{R}^{n \times n}$ and $B \in \mathbb{R}^{n \times m}$ define the system dynamics, $Q \succeq 0$, $R \succ 0$, and $Q_f \succeq 0$ are symmetric cost matrices and $q$ and $r$ are linear cost vectors. 



Equation~\eqref{eq:lqr} has a closed-form solution in the form of an affine feedback controller \cite{lewis12optimal}:
\begin{equation}\label{eq:lqrSolution} 
    u_k^* = -K_k x_k - d_k.
\end{equation}
The feedback gain $K_k$ and feedforward $d_k$ are found by solving the discrete Riccati equation backward in time, starting from $P_N = Q_f, p_N = q_f$, where $P_k, p_k$ are the quadratic and linear terms of the cost-to-go (or value) function \cite{lewis12optimal}:
\begin{equation}\label{eq:riccati}
\begin{split}
K_k &= (R+B^\intercal P_{k+1}B)^{-1} (B^\intercal P_{k+1}A)\\
d_k &= (R+B^\intercal P_{k+1}B)^{-1} (B^\intercal p_{k+1} + r_k)\\
P_k &= Q + K_k^\intercal R K_k + (A-BK_k)^\intercal P_{k+1} (A-BK_k)\\
p_k &= q_k + (A-BK_k)^\intercal (p_{k+1}-P_{k+1}Bd_k) + \\ &\qquad K_k^\intercal(Rd_k - r_k).    
\end{split}
\end{equation}

\subsection{Convex Model-Predictive Control}
Convex MPC extends the LQR formulation to admit additional convex constraints on the system states and control inputs, such as joint and torque limits, hyperplanes for obstacle avoidance, and contact constraints:
\begin{equation}\label{eq:convexMpc}
\begin{aligned}
    \min_{x_{1:N},u_{1:N-1}} \quad & J(x_{1:N},u_{1:N-1})\\
\textrm{subject to} \quad & x_{k+1} = A x_{k} + B u_{k}, \\
& x_k \in \mathcal{X}, u_k \in \mathcal{U},\\
\end{aligned}
\end{equation}
where $\mathcal{X}$ and $\mathcal{U}$ are convex sets. The convexity of this problem means that it can be solved efficiently and reliably, enabling real-time deployment in a variety of control applications including the landing of rockets~\cite{13rocketlanding}, legged locomotion~\cite{18legged}, and autonomous driving~\cite{18driving}.

When $\mathcal{X}$ and $\mathcal{U}$ can be expressed as linear 
constraints, \eqref{eq:convexMpc} is a QP, and can be put into the standard form:
\begin{equation}\label{eq:qp}
\begin{aligned}
    \min_{x \in \mathbb{R}^n} \quad &  \frac{1}{2}x^{\intercal}Px + q^{\intercal}x\\
\textrm{subject to} \quad & Ax \leq b,\\
& Cx = d.\\
\end{aligned}
\end{equation}

Further analysis, including theoretical guarantees regarding feasibility and stability can be found in~\cite{BOCCIA201414} and~\cite{MOHAMMADI201873}.



\subsection{The Alternating Direction Method of Multipliers}

The alternating direction method of multipliers (ADMM)~\cite{glowinski1975approximation,gabay1976dual,boyd2011distributed} is a popular and efficient approach for solving convex optimization problems, including QPs like \eqref{eq:qp}. We provide a very brief summary here and refer readers to \cite{boyd11admm} for more details.

Given a generic problem:
\begin{equation}\label{eq:simpleOpt}
\begin{aligned}
\min_{x} \quad & f(x)\\
\textrm{subject to} \quad & x \in \mathcal{C} , \\
\end{aligned}
\end{equation}
with $f$ and $\mathcal{C}$ convex,
we define the indicator function for the set $\mathcal{C}$:
\begin{align}
    I_{\mathcal{C}}(z) = \begin{cases}
       0& z \in \mathcal{C}  \\
       \infty& \text{otherwise}.
    \end{cases} 
    \label{eq:indicator_func}
\end{align}
We can now form the following equivalent problem by introducing the slack variable $z$:
\begin{equation}\label{eq:simpleOptTransformed}
\begin{aligned} 
\min_{x} \quad & f(x) + I_{\mathcal{C}}(z)\\
\textrm{subject to} \quad & x = z. \\
\end{aligned}
\end{equation}
The augmented Lagrangian of the transformed problem \eqref{eq:simpleOptTransformed} is as follows, where $\lambda$ is a Lagrange multiplier and $\rho$ is a scalar penalty weight:
\begin{equation}\label{eq:simpleOptTransformedAugL}
    \mathcal{L}_A(x,z,\lambda) = f(x) + I_{\mathcal{C}}(z) + \lambda^{\intercal}(x-z) + \frac{\rho}{2}||x-z||_2^2.
\end{equation}
If we alternate minimization over $x$ and $z$, rather than simultaneously minimizing over both, we arrive at the three-step ADMM iteration,
%
%
\begin{align}
    \text{primal update}: x^{+} &= \argmin_x \mathcal{L}_A(x,z,\lambda) ,\label{eq:primal_update}\\
    \text{slack update}: z^{+} &= \argmin_z \mathcal{L}_A(x^{+},z,\lambda),\label{eq:slack_update}\\
    \text{dual update}: \lambda^{+} &= \lambda + \rho (x^+ - z^+) \label{eq:dual_update},
\end{align}
the last step of which is a gradient-ascent update on the Lagrange multiplier~\cite{boyd2011distributed}. These steps can be iterated until a desired convergence tolerance is achieved. 


In the special case of a QP, each step of the ADMM algorithm becomes very simple to compute: the primal update is the solution to a linear system, and the slack update is a linear projection. ADMM-based QP solvers, like OSQP~\cite{stellato2020osqp}, have demonstrated state-of-the-art results.

\section{The TinyMPC Solver} \label{sec:algorithm}


TinyMPC trades generality for speed by exploiting the special structure of the MPC problem. Specifically, we leverage the closed-form Riccati solution to the LQR problem to compute the primal update in \eqref{eq:primal_update}. Pre-computing and caching this solution allows us to avoid online matrix factorizations and enables very fast performance while maintaining a small memory footprint. 
\vspace{40pt}

\subsection{Combining LQR and ADMM for MPC}\label{ss:combineLqrAdmm} We solve the following problem, introducing slack variables as in~\eqref{eq:simpleOptTransformedAugL} and transforming~\eqref{eq:convexMpc} into the following:
\vspace{-5pt}
\begin{equation}\label{eq:lqr_admm}
\begin{aligned}
    \min_{{\substack{x_{1:N}, {z}_{1:N},\\{\lambda }_{1:N}, u_{1:N-1},\\{w}_{1:N-1}, {\mu}_{1:N-1}\\}}} & \begin{split} \! \\
        \mathcal{L}_A(\cdot) = &~J(x_{1:N},u_{1:N-1}) + \\
    &I_{\mathcal{X}}({z}_{1:N}) + I_{\mathcal{U}}({w}_{1:N-1}) +
    \end{split}\\
    \quad \; \sum_{k=1}^{N} \frac{\rho}{2}(x_k&-{z}_k)^{\intercal}(x_k-{z}_k) + {\lambda}_{k}^{\intercal}(x_k-{z}_k) + \\
    \quad \; \sum_{k=1}^{N-1} \frac{\rho}{2}(u_k&-{w}_k)^{\intercal}(u_k-{w}_k) + {\mu}_{k}^{\intercal}(u_k-{w}_k)\\
\textrm{subject to:} \quad & \; x_{k+1} = A x_{k} + B u_{k},\\
\end{aligned}
\end{equation}
\vspace{5pt}

\noindent where $z$, $w$, $\lambda$, and $\mu$ are the state slack, input slack, state dual, and input dual variables over the entire horizon. State and input constraints are enforced through the indicator functions $I_{\mathcal{X}}$ and $I_{\mathcal{U}}$. We use the ADMM algorithm \eqref{eq:primal_update}, \eqref{eq:slack_update}, \eqref{eq:dual_update} to solve this optimal control problem. The primal update for~\eqref{eq:lqr_admm} becomes an equality-constrained QP: 
\begin{equation}\label{eq:lqr_primal}
\begin{aligned}
    \min_{x_{1:N},u_{1:N-1}} \quad & \frac{1}{2}x_N^{\intercal} \tilde{Q}_f x_N + \tilde{q}_f^{\intercal}x_N + \\
    \sum_{k = 1}^{N-1} & \frac{1}{2}x_k^{\intercal}\tilde{Q}x_k + \tilde{q}_k^{\intercal}x_k
    + \frac{1}{2}u_k^{\intercal}\tilde{R}u_k + \tilde{r}^{\intercal}u_k\\
\textrm{subject to} \quad & x_{k+1} = A x_{k} + B u_{k} ,\\
\end{aligned}
\end{equation}
where
\begin{equation}\label{eq:lin_quad_updated_cost}
    \begin{aligned}
        &\tilde{Q}_f = Q_f + \rho I, && \quad 
        \tilde{q}_f = q_f + \lambda_N - \rho z_N,\\
        &\tilde{Q} = Q + \rho I, && \quad
        \tilde{q}_k = q_k + \lambda_k - \rho z_k,\\
        &\tilde{R} = R + \rho I, &&\quad
        \tilde{r}_k = r_k + \mu_k - \rho w_k.\\
    \end{aligned}
\end{equation}

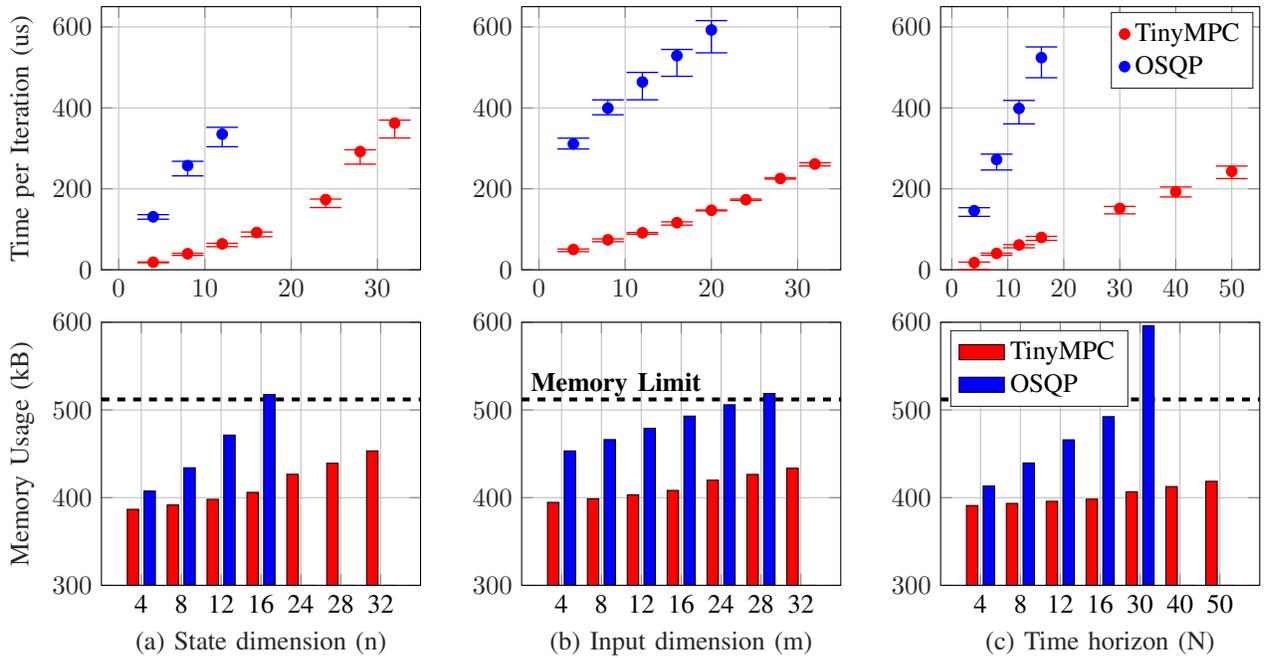
\begin{figure*}[!ht]
    \centering
    \vspace{5pt}
    \setlength{\figW}{0.75\paperwidth}
    \setlength{\figH}{3.5cm}
%
%
\begin{tikzpicture}

\begin{axis}[%
width=0.262\figW,
height=\figH,
at={(0\figW,0\figH)},
scale only axis,
xmin=-2,
xmax=35,
xtick={ 0, 10, 20, 30},
ymin=0,
ymax=650,
ylabel style={font=\color{white!15!black}},
ylabel={Time per Iteration (us)},
axis background/.style={fill=white},
xmajorgrids,
ymajorgrids
]
\addplot [color=red, only marks, mark=*, mark options={solid, fill=red}, forget plot]
 plot [error bars/.cd, y dir=both, y explicit, error bar style={line width=0.5pt}, error mark options={line width=0.5pt, mark size=6.0pt, rotate=90}]
 table[row sep=crcr, y error plus index=2, y error minus index=3]{%
4	18.695354817298	0.471311849368632	1.36436890180508\\
8	39.6127372505253	1.13726274947468	4.04475479438497\\
12	64.0736926141408	1.17630738585925	6.91937540654737\\
16	92.0985580056882	1.15144199431177	10.2885487378476\\
24	173.451549849501	1.29845015049881	19.7132998495012\\
28	291.917005697432	4.58299430256767	30.8527556974323\\
32	362.425548236139	7.32445176386125	36.8320482361387\\
};
\addplot [color=blue, only marks, mark=*, mark options={solid, fill=blue}, forget plot]
 plot [error bars/.cd, y dir=both, y explicit, error bar style={line width=0.5pt}, error mark options={line width=0.5pt, mark size=6.0pt, rotate=90}]
 table[row sep=crcr, y error plus index=2, y error minus index=3]{%
4	130.908593723517	5.29140627648263	5.90859372351736\\
8	257.438618143888	10.5613818561123	25.4386181438877\\
12	335.131999144655	16.8680008553454	31.1319991446546\\
};
\end{axis}

\begin{axis}[%
width=0.262\figW,
height=\figH,
at={(0.345\figW,0\figH)},
scale only axis,
xmin=-2,
xmax=35,
xtick={ 0, 10, 20, 30},
ymin=0,
ymax=650,
axis background/.style={fill=white},
xmajorgrids,
ymajorgrids
]
\addplot [color=red, only marks, mark=*, mark options={solid, fill=red}, forget plot]
 plot [error bars/.cd, y dir=both, y explicit, error bar style={line width=0.5pt}, error mark options={line width=0.5pt, mark size=6.0pt, rotate=90}]
 table[row sep=crcr, y error plus index=2, y error minus index=3]{%
4	50.1700434575046	1.07995654249536	5.39064299632093\\
8	74.0256479797681	2.17435202023191	4.60374492770346\\
12	91.3353453587658	0.664654641234208	3.38689175051837\\
16	116.305320416426	2.09467958357433	5.92070503181029\\
20	147.028859649123	0.771140350877175	1.52885964912284\\
24	173.2325	1.51749999999998	1.83250000000001\\
28	225.463947368421	1.53605263157925	1.46394736842075\\
32	261.234912280702	3.26508771929838	4.73491228070162\\
};
\addplot [color=blue, only marks, mark=*, mark options={solid, fill=blue}, forget plot]
 plot [error bars/.cd, y dir=both, y explicit, error bar style={line width=0.5pt}, error mark options={line width=0.5pt, mark size=6.0pt, rotate=90}]
 table[row sep=crcr, y error plus index=2, y error minus index=3]{%
4	310.891688512583	14.3083114874169	12.3916885125831\\
8	399.265844889609	20.5341551103911	16.265844889609\\
12	463.923975309896	23.6760246901041	43.9239753098959\\
16	529.004306526806	15.3956934731935	51.0043065268064\\
20	592.453026315789	23.1469736842106	56.4530263157894\\
};
\end{axis}

\begin{axis}[%
width=0.262\figW,
height=\figH,
at={(0.689\figW,0\figH)},
scale only axis,
xmin=-2,
xmax=55,
xtick={ 0, 10, 20, 30, 40, 50},
ymin=0,
ymax=650,
axis background/.style={fill=white},
xmajorgrids,
ymajorgrids,
legend style={legend cell align=left, align=left, draw=white!15!black}
]
\addplot [color=red, only marks, mark=*, mark options={solid, fill=red}]
 plot [error bars/.cd, y dir=both, y explicit, error bar style={line width=0.5pt}, error mark options={line width=0.5pt, mark size=6.0pt, rotate=90}]
 table[row sep=crcr, y error plus index=2, y error minus index=3]{%
4	17.517335096906	1.48266490309397	17.4855481553253\\
8	40.447120416503	0.552879583497038	4.70346767285022\\
12	61.3636923534013	0.636307646598674	7.06288752442748\\
16	79.7578124521806	2.74218754781943	7.13096060032872\\
30	151.588551176492	4.91144882350815	13.412837690202\\
40	192.849150479655	11.6508495203448	12.9361495021381\\
50	243.120842778169	13.1291572218312	17.7478505116608\\
};
\addlegendentry{TinyMPC}

\addplot [color=blue, only marks, mark=*, mark options={solid, fill=blue}]
 plot [error bars/.cd, y dir=both, y explicit, error bar style={line width=0.5pt}, error mark options={line width=0.5pt, mark size=6.0pt, rotate=90}]
 table[row sep=crcr, y error plus index=2, y error minus index=3]{%
4	145.999297905859	7.40070209414094	13.9992979058591\\
8	272.411132742371	13.3888672576288	25.9111327423713\\
12	398.603953447718	19.7960465522821	38.1039534477179\\
16	524.132196545405	26.4678034545947	49.6321965454053\\
};
\addlegendentry{OSQP}

\end{axis}
\end{tikzpicture}%
%
%
\begin{tikzpicture}

\begin{axis}[%
width=0.262\figW,
height=\figH,
at={(0\figW,0\figH)},
scale only axis,
bar shift auto,
log origin=infty,
xmin=0,
xmax=8,
xtick={1,2,3,4,5,6,7},
xticklabels={{4},{8},{12},{16},{24},{28},{32}},
xticklabel style={rotate=0, font = {\mdseries}},
xlabel style={font=\color{white!15!black}},
xlabel={(a) State dimension (n)},
ymin=300,
ymax=600,
ylabel style={font=\color{white!15!black}},
ylabel={Memory Usage (kB)},
axis background/.style={fill=white},
xmajorgrids,
ymajorgrids
]
\addplot [color=black, dashed, line width=1.5pt, forget plot]
  table[row sep=crcr]{%
0	512\\
32	512\\
};
\addplot[ybar, bar width=0.286, fill=red, draw=black, area legend] table[row sep=crcr] {%
1	386.656\\
2	391.584\\
3	398.016\\
4	406.016\\
5	426.592\\
6	439.168\\
7	453.312\\
};
\addplot[ybar, bar width=0.286, fill=blue, draw=black, area legend] table[row sep=crcr] {%
1	407.52\\
2	434.032\\
3	471.32\\
4	517.568\\
5	0\\
6	0\\
7	0\\
};
\end{axis}

\begin{axis}[%
width=0.262\figW,
height=\figH,
at={(0.345\figW,0\figH)},
scale only axis,
bar shift auto,
log origin=infty,
xmin=0,
xmax=8,
xtick={1,2,3,4,5,6,7},
xticklabels={{4},{8},{12},{16},{24},{28},{32}},
xticklabel style={rotate=0, font = {\mdseries}},
xlabel style={font=\color{white!15!black}},
xlabel={(b) Input dimension (m)},
ymin=300,
ymax=600,
axis background/.style={fill=white},
xmajorgrids,
ymajorgrids
]
\addplot [color=black, dashed, line width=1.5pt, forget plot]
  table[row sep=crcr]{%
0	512\\
32	512\\
};
\node [anchor=south west,font=\fontsize{10}{10}\selectfont\bfseries] at (axis cs: 0, 505) {Memory Limit};
\addplot[ybar, bar width=0.286, fill=red, draw=black, area legend] table[row sep=crcr] {%
1	394.592\\
2	398.624\\
3	403.168\\
4	408.224\\
5	419.872\\
6	426.464\\
7	433.568\\
};
\addplot[ybar, bar width=0.286, fill=blue, draw=black, area legend] table[row sep=crcr] {%
1	453.28\\
2	466.176\\
3	479.072\\
4	492.992\\
5	505.888\\
6	518.784\\
7	0\\
};
\end{axis}

\begin{axis}[%
width=0.262\figW,
height=\figH,
at={(0.689\figW,0\figH)},
scale only axis,
bar shift auto,
log origin=infty,
xmin=0,
xmax=8,
xtick={1,2,3,4,5,6,7},
xticklabels={{4},{8},{12},{16},{30},{40},{50}},
xticklabel style={rotate=0, font = {\mdseries}},
xlabel style={font=\color{white!15!black}},
xlabel={(c) Time horizon (N)},
ymin=300,
ymax=600,
axis background/.style={fill=white},
xmajorgrids,
ymajorgrids,
legend style={at={(0.03,0.97)}, anchor=north west, legend cell align=left, align=left, draw=white!15!black}
]
\addplot [color=black, dashed, line width=1.5pt, forget plot]
  table[row sep=crcr]{%
0	512\\
32	512\\
};
\addplot[ybar, bar width=0.286, fill=red, draw=black, area legend] table[row sep=crcr] {%
1	391.008\\
2	393.408\\
3	395.808\\
4	398.208\\
5	406.592\\
6	412.608\\
7	418.592\\
};
\addlegendentry{TinyMPC}

\addplot[ybar, bar width=0.286, fill=blue, draw=black, area legend] table[row sep=crcr] {%
1	413.216\\
2	439.584\\
3	465.952\\
4	492.32\\
5	596.144\\
6	0\\
7	0\\
};
\addlegendentry{OSQP}

\end{axis}
\end{tikzpicture}%
    \caption{Comparison of average iteration times (top) and memory usage (bottom) for OSQP and TinyMPC on randomly generated trajectory tracking problems on a Teensy 4.1 development board (ARM Cortex-M7 running at 600 MHz with 32-bit floating point support, 7.75 MB of flash, and 512 kB of tightly coupled RAM). Error bars show the maximum and minimum time per iteration over all MPC steps executed for a given problem. In (a), the input dimension and time horizon are held constant at $m=4$ and $N=10$ while the state dimension $n$ varies from 4 to 32. In (b), $n=10$ and $N=10$ while the $m$ varies from 4 to 32. In (c), $n=10, m=4$ and $N$ varies from 4 to 50. The dotted black line indicates the memory limit of the Teensy 4.1.}
    \label{fig:mcu_comparison}
    \vspace{-8pt}
\end{figure*}

We reformulate \eqref{eq:lin_quad_updated_cost} and introduce the scaled dual variables $y$ and $g$ for convenience~\cite{boyd2011distributed}:
\begin{equation}\label{eq:dual_scale}
    \begin{aligned}
        &\tilde{q}_f = q_f + \rho(\lambda_N / \rho -  z_N) = q_f + \rho(y_N -  z_N),\\
        &\tilde{q}_k = q_k + \rho(\lambda_k / \rho - z_k) = q_k + \rho(y_k - z_k),\\
        &\tilde{r}_k = r_k + \rho(\mu_k / \rho -  w_k) = r_k + \rho(g_k -  w_k).\\
    \end{aligned}
\end{equation}
We observe that, because \eqref{eq:lqr_primal} exhibits the same LQR problem structure as in \eqref{eq:lqr}, it can be solved efficiently with the Riccati recursion in \eqref{eq:riccati}. 
The slack update for \eqref{eq:lqr_admm} becomes a simple linear projection onto the feasible set:
\begin{equation}\label{eq:mpc_slack}
    \begin{aligned}
    z^{+}_k &= \proj _\mathcal{X} (x^+_k+y_k), \\
    w^{+}_k &= \proj _\mathcal{U} (u^+_k+g_k),
    \end{aligned}
\end{equation}
where the superscript denotes the variable at the subsequent ADMM iteration. The dual update for~\eqref{eq:lqr_admm} becomes:
\begin{equation}
    \begin{aligned}\label{eq:mpc_dual}
    y^{+}_k &= y_k + x^+_k -  z^+_k,\\
    g^{+}_k &= g_k + u^+_k -  w^+_k .\\
    \end{aligned}
\end{equation}
Finally, the algorithm terminates when the primal and dual residuals are within a set tolerance.
\newpage

\subsection{Pre-Computation}\label{ss:precompute}

Solving the linear system in each primal update is the most expensive step in each ADMM iteration. In our case, this is the solution to the Riccati equation, which has properties we can leverage to significantly reduce computation and memory usage. Given a long enough horizon, the Riccati recursion \eqref{eq:riccati} converges to the constant solution of the infinite-horizon LQR problem \cite{lewis12optimal}. Thus, we pre-compute a single LQR gain matrix $K_{\text{inf}}$ and cost-to-go Hessian $P_{\text{inf}}$. We then cache the following matrices from \eqref{eq:riccati}:
\begin{equation}\label{eq:cache}
    \begin{aligned}
        C_1 &= (R + B^\intercal P_\text{inf}B)^{-1}, \\
        C_2 &= (A-BK_\text{inf})^\intercal. \\
    \end{aligned}
\end{equation}
A careful analysis of the Riccati equation then reveals that only the linear terms need to be updated as part of the ADMM iteration:
\begin{equation}\label{eq:fast_riccati}
    \begin{aligned}
        d_k &= C_1(B^\intercal p_{k+1} + r_k), \\
        p_k &= q_k + C_2 p_{k+1} - K_\text{inf}^\intercal r_k. \\
    \end{aligned}
\end{equation}
As a result, we completely avoid online matrix factorization and only compute matrix-vector products. We also dramatically reduce memory footprint by only storing a few vectors at each time step.




\subsection{Penalty Scaling}\label{ss:scaling}

ADMM is sensitive to the value of the penalty term $\rho$ in \eqref{eq:simpleOptTransformedAugL}. Solvers like OSQP~\cite{stellato2020osqp} overcome this issue by adaptively scaling $\rho$. However, this requires performing additional matrix factorizations. To avoid this, we pre-compute and cache a set of matrices corresponding to several values of $\rho$. Online, we switch between these cached matrices based on the values of the primal and dual residual values in a scheme adapted from OSQP. 
The resulting TinyMPC algorithm is summarized in Algorithm \ref{alg:tinympc_alg_words}. 

\begin{algorithm}[!h]
\caption{TinyMPC}\label{alg:tinympc_alg_words}
    \begin{algorithmic}
    \Function{tiny\underline{{ }{ }}solve}{input}
        \While {\text{not converged}}
        \State \texttt{//Primal update}
        \State $ p_{1:N-1}, d_{1:N-1} \gets \text{Backward pass via \eqref{eq:fast_riccati}}$
        \State $x_{1:N}, u_{1:N-1} \gets \text{Forward pass via \eqref{eq:lqrSolution}}$
        
        \State \texttt{//Slack update}
        \State $z_{1:N}, w_{1:N-1} \gets \text{Project to feasible set \eqref{eq:mpc_slack}}$
        
        \State \texttt{//Dual update}
        \State $y_{1:N}, g_{1:N-1}  \gets \text{Gradient ascent \eqref{eq:mpc_dual}}$
        \State $q_{1:N}, r_{1:N-1}, p_N \gets \text{Update linear cost terms}$
        \EndWhile
        
        \Return $x_{1:N}, u_{1:N-1}$
    \EndFunction
    \end{algorithmic}
\end{algorithm}
\vspace{-10pt}

\section{Experiments} \label{sec:results}
\begin{figure}[h]
    \vspace{5pt}
    \centering
    \setlength{\figW}{0.8\columnwidth}
    \setlength{\figH}{15cm}
    \input{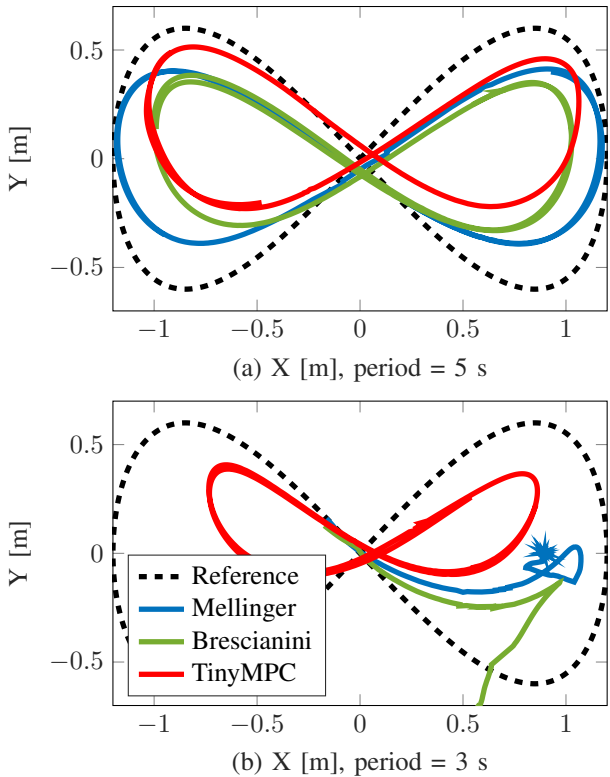}
    \caption{Figure-eight tracking at low speed (top) and high speed (bottom) comparing TinyMPC with the two most performant controllers available on the Crazyflie. For slower trajectories, all three controllers resulted in similar performance. For faster trajectories, only TinyMPC was capable of maintaining tracking without crashing. The maximum velocity and attitude deviation from hover with TinyMPC reached 1.5 m/s and 20$^\circ$, respectively.}
    \label{fig:fig8_pos}
\end{figure}

\begin{figure}[h]
    \vspace{5pt}
    \centering
    \setlength{\figW}{0.8\columnwidth}
    \setlength{\figH}{5cm}
    \input{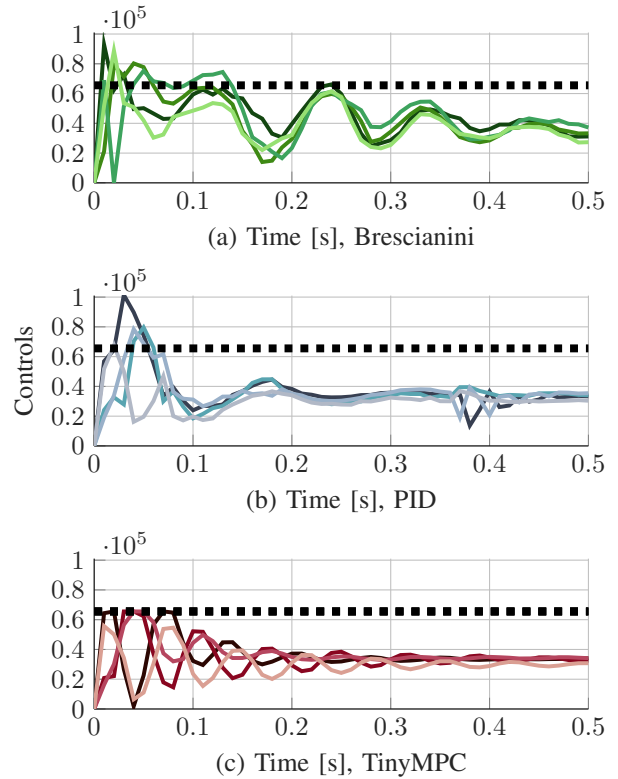}
    \caption{Control trajectories during the Extreme Initial Poses experiment. Four pre-clipped PWM motor commands are shown for each controller. The black dotted line denotes the thrust limit, from 0 to 65535 PWM value. Among the three successful controllers, only TinyMPC could reason about control feasibility, exhibiting the maneuver shown in Fig.~\ref{fig:avoid2} (bottom).}
    \label{fig:recovery_controls}
\end{figure}


We evaluate TinyMPC through two sets of experiments: first, we benchmark our solver against the state-of-the-art OSQP~\cite{stellato2020osqp} solver on a representative microcontroller, demonstrating improved computational speed and reduced memory footprint. We then test the efficacy of our solver on a resource-constrained nano-quadrotor platform, the Crazyflie 2.1. We show that TinyMPC enables the Crazyflie to track aggressive reference trajectories while satisfying control limits and time-varying state constraints.


\subsection{Microcontroller Benchmarks}\label{ss:benchmark}

As shown in Fig.~\ref{fig:mcu_comparison}, we first compare TinyMPC and OSQP on random linear MPC problems while varying the state and input dimensions, as well as the horizon length.

\subsubsection{Methodology} Experiments were performed on a Teensy 4.1~\cite{teensy41} development board, 
which has an ARM Cortex-M7 microcontroller operating at 600 MHz, 7.75 MB of flash memory, and 512 kB of RAM.
TinyMPC is implemented in C++ using the Eigen matrix library \cite{eigenweb}. We used OSQP's code-generation feature to generate a C implementation of each problem to run on the microcontroller. 
Objective tolerances were set to $10^{-3}$ and constraint tolerances to $10^{-4}$. The maximum number of iterations for both solvers was set to 4000, and both utilized warm starting. OSQP's solution polishing was disabled to decrease solve time. Other parameters were set to equivalent values wherever possible.

Dynamics models $A$ and $B$ were randomly generated and checked to ensure controllability for all values of state dimension $n$, input dimension $m$, and time horizon $N$. The control input was constrained within fixed bounds over the entire horizon. 
During the microcontroller tests, noise was added to mimic imperfect state estimation. 
The largest problem instance involved 696 decision variables, 490 linear equality constraints, and 392 linear inequality constraints.

\subsubsection{Evaluation} 
Fig.~\ref{fig:mcu_comparison} shows the average execution times for both solvers, in which TinyMPC exhibits a maximum speed-up of 8.85x over OSQP. This speed-up allows TinyMPC to perform real-time trajectory tracking while handling input constraints. OSQP also quickly exceeded the memory limitations of the MCU, while TinyMPC was able to scale to much larger problem sizes. For example, for a fixed input dimension of $m=4$ and time horizon of $N=10$ (Fig.~\ref{fig:mcu_comparison}a), OSQP exceeded the 512 kB memory limit of the Teensy at a state dimension of only $n=16$, while TinyMPC only used around 400 kB at a state dimension of $n=32$.

\subsection{Hardware Experiments}\label{ss:demos}

We demonstrate the efficacy of our solver for real-time execution of dynamic control tasks on a resource-constrained Crazyflie 2.1 quadrotor. We present three experiments: 1)~figure-eight trajectory tracking at slow and fast speeds, 2)~recovery from extreme initial attitudes, and 3)~dynamic obstacle avoidance through online updating of state constraints.

\subsubsection{Methodology} The Crazyflie 2.1 is a 27 gram quadrotor. Its primary MCU is an ARM Cortex-M4 (STM32F405) clocked at 168 MHz with 192 kB of SRAM and 1 MB of flash. OSQP could not fit within the memory available on this MCU, making it impossible to be used as an MPC baseline. Instead, we compare against the four controllers included with the Crazyflie firmware: Cascaded PID~\cite{cf_pid}, Mellinger~\cite{mel11mel}, INDI~\cite{smeur16indi}, and Brescianini~\cite{bre2013bre}. These are reactive controllers that often clip the control input to meet hardware constraints. 

All experiments shown were performed in an OptiTrack motion-capture environment sending pose data to the Crazyflie at 100 Hz.
TinyMPC ran at 500 Hz with a horizon length of $N=15$ for the figure-eight tracking task and the attitude-recovery task. 
For the obstacle-avoidance task, we sent the location of the end of a stick to the Crazyflie using the onboard radio. Additionally, we reduced the MPC frequency to 100 Hz and increased $N$ to 20.

In all experiments, we linearized the quadrotor's 6-DOF dynamics about a hover and represented its attitude with a quaternion using the formulation in~\cite{jackson21planning}. This problem has state dimension $n=12$ and $m=4$ representing the quadrotor's full state and PWM motor commands. The largest problem was in the dynamic obstacle avoidance scenario, which was solved onboard at high frequency and consisted of 316 decision variables, 248 linear equality constraints, and 172 linear inequality constraints.


\subsubsection{Evaluation––Figure-Eight Trajectory Tracking}
We compare the tracking performance of TinyMPC and other controllers with a figure-eight trajectory, as shown in Fig.~\ref{fig:fig8_pos}. For the faster trajectory, the maximum velocity and attitude deviation reached 1.5 m/s and 20$^\circ$, respectively. Only TinyMPC could track the entire reference while respecting actuator limits, while the Mellinger and Brescianini controllers crashed almost immediately. 
TinyMPC converged at all steps within a maximum of 7 iterations and under the allotted 2 ms solve time defined by the 500 Hz control frequency.

\subsubsection{Evaluation––Extreme Initial Poses}
Fig.~\ref{fig:avoid2} (bottom) shows the performance of the Crazyflie when initialized with a 90$^\circ$ attitude error. TinyMPC displayed the best recovery performance with a maximum position error of 23~cm while respecting the input limits. The PID and Brescianini controllers achieved maximum errors of 40 cm and 65~cm, respectively, while violating input limits (Fig. \ref{fig:recovery_controls}). The other controllers, INDI and Mellinger, failed to stabilize the quadrotor, causing it to crash. 



\subsubsection{Evaluation––Dynamic Obstacle Avoidance}
We demonstrate TinyMPC's ability to handle time-varying state constraints by avoiding a moving stick (Fig.~\ref{fig:avoid2} top). These experiments are more challenging because the constraints arbitrarily switch between inactive and active, requiring far more iterations to solve to convergence. The obstacle sphere was re-linearized about its updated position at each MPC step, allowing the drone to avoid the unplanned movements of the swinging stick. 
As illustrated, the quadrotor could move freely in space to avoid the dynamic obstacle and come back safely to the hovering position. As an additional challenge, we added a constraint such that the quadrotor must stay within a vertical plane defined by $x=0$. The Crazyflie deviated a maximum of approximately 5 cm from this constraint plane while successfully avoiding the dynamic obstacle.

\section{Conclusion and Future Work} \label{sec:conclusion}
We introduce TinyMPC, a model-predictive control solver for resource-constrained embedded systems. TinyMPC uses ADMM to handle state and input constraints while leveraging the structure of the MPC problem and insights from LQR to reduce memory footprint and speed up online execution compared to existing state-of-the-art solvers like OSQP. We demonstrated TinyMPC's practical performance on a Crazyflie nano-quadrotor performing highly dynamic tasks with input and obstacle constraints.



Several directions for future work remain. First, it should be straightforward to extend TinyMPC to handle second-order cone constraints, which are useful in many MPC applications for modeling thrust and friction. We also plan to further reduce TinyMPC's hardware requirements by developing a fixed-point version, since many small microcontrollers lack hardware floating-point support.
Finally, to ease deployment and adoption, we plan to develop a code-generation wrapper for TinyMPC in a high-level language such as Julia or Python, similar to OSQP and CVXGEN.

For more information and to get started using our open-source solver, we recommend users visit our website, {\color{blue}{\url{https://tinympc.org}}}.

\section{Acknowledgments} \label{sec:ack}
The authors would like to thank Brian Jackson for insightful discussions and Professor Mark Bedillion for providing us with extra Crazyflies.

\bibliographystyle{bib/IEEEtran}
\bibliography{bib/IEEEabrv,bib/agile.bib}

\end{document}